\documentclass[runningheads]{llncs}
\usepackage{graphicx}
\usepackage{comment}
\usepackage{amsmath,amssymb} 
\usepackage{color}

\usepackage{comment}
\usepackage{boldline}
\usepackage{tabularx}
\usepackage{capt-of}
\usepackage{mathtools}
\usepackage{array,multirow,float}
\usepackage{bm}
\usepackage[caption=false,font=footnotesize]{subfig}
\usepackage{algorithm} 
\usepackage{algpseudocode}
\usepackage[final]{pdfpages}

\usepackage{booktabs}

\newcolumntype{L}{>{$}l<{$}}
\newcolumntype{C}{>{$}c<{$}}
\newcolumntype{R}{>{$}r<{$}}

\newcommand{\reals}{{\mathbb{R}}}

\newcommand{\identityf}[1]{\mathbf 1_{\{#1\}}}

\usepackage[dvipsnames]{xcolor}

\newcommand{\calC}{{\cal C}}

\newcommand{\calL}{{\cal L}}

\usepackage[pagebackref=true,breaklinks=true,letterpaper=true,colorlinks,bookmarks=false]{hyperref}

\usepackage{wrapfig}

\begin{document}
\pagestyle{headings}
\mainmatter
\def\ECCVSubNumber{5625}  

\title{DBQ: A Differentiable Branch Quantizer for Lightweight Deep Neural Networks} 

\titlerunning{A Differentiable Branch Quantizer for Lightweight Deep Neural Networks}
%
\author{Hassan Dbouk\inst{1,2}\thanks{Work done while at Kilby Labs.} \and
Hetul Sanghvi \inst{2}\and
Mahesh Mehendale \inst{2} \and Naresh Shanbhag \inst{1}}
\authorrunning{H. Dbouk et al.}
%
\institute{Dept. of Electrical and Computer Engineering, University of Illinois at Urbana-Champaign, Urbana, USA \\
\email{\{hdbouk2, shanbhag\}@illinois.edu}\and Kilby Labs, Texas Instruments Inc, Dallas, USA \\ \email{\{hetul, m-mehendale\}@ti.com}}
\maketitle

\begin{abstract}
Deep neural networks have achieved state-of-the art performance on various computer vision tasks. However, their deployment on resource-constrained devices has been hindered due to their high computational and storage complexity. While various complexity reduction techniques, such as lightweight network architecture design and parameter quantization, have been successful in reducing the cost of implementing these networks, these methods have often been considered orthogonal. In reality, existing quantization techniques fail to replicate their success on lightweight architectures such as MobileNet. To this end, we present a novel fully differentiable non-uniform quantizer that can be seamlessly mapped onto efficient ternary-based dot product engines. We conduct comprehensive experiments on CIFAR-10, ImageNet, and Visual Wake Words datasets. The proposed quantizer (DBQ) successfully tackles the daunting task of aggressively quantizing lightweight networks such as MobileNetV1, MobileNetV2, and ShuffleNetV2. DBQ achieves state-of-the art results with minimal training overhead and provides the best (pareto-optimal) accuracy-complexity trade-off. 
\end{abstract}

\section{Introduction}

Deep neural networks (DNNs) have achieved state-of-the art accuracy on various computer vision tasks such as image classification \cite{krizhevsky2012imagenet,he2016deep} but at the expense of extremely high computational and storage complexity, e.g., ResNet-18 \cite{he2016deep} needs $\sim 10^{12}$ 1-b full adders (FAs) and $3.74\times 10^{8}$-bits of activation and weight storage to achieve an accuracy of 70\% on the ImageNet dataset. These high computational and storage costs inhibit the deployment of such DNNs on resource-constrained Edge devices. As a result, there is much interest in designing low-complexity DNNs without compromising their accuracy.

There are two distinct approaches for reducing DNN complexity: 1) model compression \cite{han2015learning} and quantization \cite{hubara2016binarized,rastegari2016xnor} of complex networks, and 2) the design of lightweight networks from scratch, e.g., MobileNet \cite{howard2017mobilenets,sandler2018mobilenetv2}. 

Model compression and quantization methods rely on the intrinsic over parameterization in complex networks to reduce their complexity. Such methods have proved to be very effective in reducing network complexity with negligible impact on its accuracy, e.g., ternary quantization of ResNet-18 weights \cite{yang2019quantization} reduces its computational and storage complexity by $88\%$ and $74\%$, respectively, at the expense of a drop in accuracy from $70.3\%$ to $69.1\%$.

In the second approach, the design of lightweight networks such as MobileNet \cite{howard2017mobilenets,sandler2018mobilenetv2}, SqueezeNet \cite{iandola2016squeezenet}, ShuffleNet \cite{zhang2018shufflenet}, ConDenseNet \cite{huang2018condensenet} have also shown tremendous success. Such networks exploit algorithmic properties such as factorizability of convolutions and utilize either $1\times 1$ convolutions (SqueezeNet), grouped convolutions (ShuffleNet, ConDenseNet), or both (MobileNet). For example, MobileNetV1 \cite{howard2017mobilenets} achieves comparable (or even higher) accuracy than its ResNet-18 floating-point (FP) counterpart but at a computational and storage complexity that are $3\times$ and $7\times$ lower, respectively.

\begin{figure}[t]
    \begin{center}
    \includegraphics[width=0.75\columnwidth]{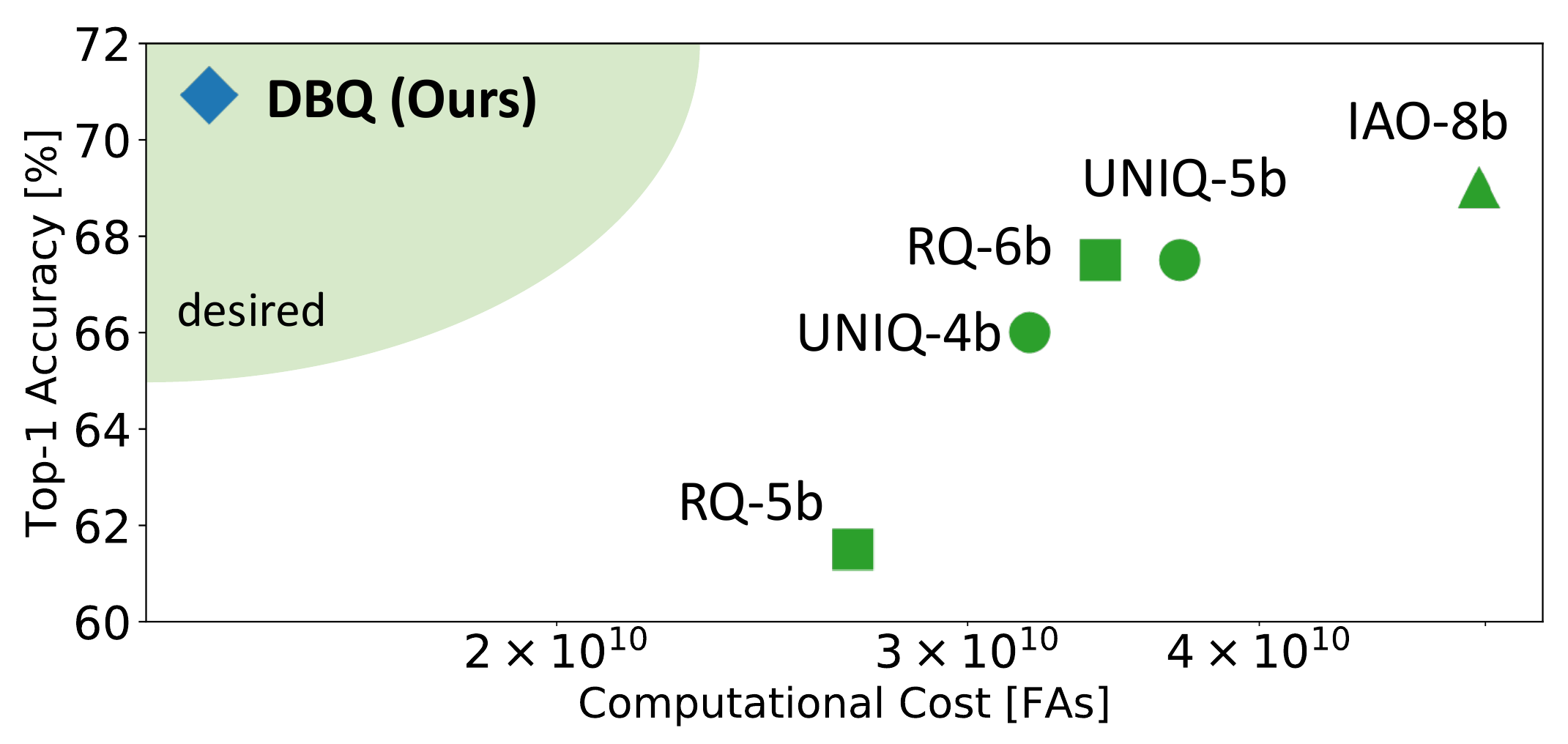}%
    \end{center}
    \caption{The Top-1 accuracy on ImageNet vs. computational cost for MobileNetV1 achieved by state-of-the-art quantization methods (RQ \cite{louizos2018relaxed}, UNIQ \cite{baskin2018uniq}, and IAO \cite{jacob2018quantization}). Our proposed method DBQ simultaneously achieves the highest accuracy and the lowest complexity.}%
    \label{fig:motivation}%
\end{figure}

In contrast, not much work has been done in model compression or quantization of lightweight networks and for a good reason -- such networks are already irredundant leaving much less room for complexity reduction. Existing works \cite{louizos2018relaxed,wang2019haq,baskin2018uniq,jacob2018quantization,sheng2018quantization} that quantize lightweight networks use fixed-point quantization with relatively high bitwidths (see Fig.~\ref{fig:motivation}) which offer limited reductions in complexity. In contrast, aggressive quantization schemes such as binarization \cite{hubara2016binarized,rastegari2016xnor} or ternarization \cite{zhu2016trained,li2016ternary} have been benchmarked on over-parameterized networks. In fact, ternarizing MobileNetV1 leads to a catastrophic drop in accuracy from $72.12\%$ to $66.45\%$ on ImageNet as we show in Section~\ref{sec:imagenet}. In order to improve the performance of ternarized models while leveraging the simplicity of ternary-based arithmetic, one can construct a non-uniform quantizer as linear combinations of ternary values. Such formulation has already been proposed in the context of binarized neural networks \cite{zhang2018lq,lin2017towards}, however the training algorithm involved is: 1) extremely inefficient to implement; 2) can lead to sub-optimal results due to gradient mismatch issues; and 3) has only been benchmarked on over-parameterized networks.

To this end, our work is the \textit{first} to tackle the daunting task of aggressively quantizing lightweight networks, such as MobileNetV1 \cite{howard2017mobilenets}, MobileNetV2 \cite{sandler2018mobilenetv2}, and ShuffleNetV2 \cite{ma2018shufflenet} using multiple ternary branches. We propose an efficient and fully differentiable multiple ternary branch quantization algorithm (DBQ). For MobileNetV1 on ImageNet, DBQ achieves an accuracy $2\%$ higher than state-of-the art quantization methods with a complexity that is $3.5\times$ lower as shown in Fig.~\ref{fig:motivation}. This represents an overall reduction of $24.5\times$ compared to FP with a $1.2\%$ drop in accuracy.

Specifically, our contributions are:
\begin{enumerate}
    \item We are the \textit{first} to successfully ternarize lightweight networks (MobileNetV1, MobileNetV2, ShuffleNetV2) on ImageNet. This result is achieved by using DBQ with two ternary branches.
    \item We present the \textit{first fully differentiable} branched quantization algorithm (DBQ) for DNNs requiring minimal training overhead.
    \item We show that DBQ outperforms state-of-the art methods in both accuracy and computational cost. Compared to state-of-the art quantization method RQ \cite{louizos2018relaxed}, DBQ drastically improves the Top-1 accuracy of MobileNetV1 on ImageNet from $61.50\%$ to $70.92\%$ at iso-model size accompanied by a $19\%$ reduction in computational complexity.
    \item For lightweight networks tackling real world applications, we show that DBQ with two ternary branches offers the best (pareto-optimal) accuracy-complexity trade-off compared to using one ternary branch with higher number of channels, at iso-model size.
\end{enumerate}


\begin{wraptable}{R}{0.45\columnwidth}
    \begin{center}
    \resizebox{0.45\columnwidth}{!}{%
    \renewcommand{\arraystretch}{1.2}
    \begin{tabular}{c c c c }
    \clineB{1-4}{2.5}
    \textbf{Layer Type} & \textbf{Mults} [\%]& \textbf{Adds} [\%]& \textbf{Params} [\%] \\\clineB{1-4}{2.5}
    FL & $1.89$ & $1.83$ & $0.02$ \\\hline
    DW & $3.03$ & $2.72$ & $1.05$ \\\hline
    PW & $\mathbf{94.02}$ & $\mathbf{94.37}$ & $\mathbf{74.19}$ \\\hline
    FC & $0.18$ & $0.18$ & $24.22$ \\\hline
    PL & $0$ & $0.01$ & $0$ \\\hline
    BN & $0.88$ & $0.89$ & $0.52$ \\\hline
    \end{tabular}
    }
    \end{center}
    \caption{The number of multiplications, additions, and parameters required by each layer type: first layer (FL), depthwise (DW), pointwise (PW), fully connected (FC), pooling layer (PL), and batch normalization (BN), for a single inference using MobileNetV1.}
    \label{tab:mobilenet-stats}
\end{wraptable} 
\section{Related Work}

Reducing DNN complexity via quantization has been an active area of research over the past few years. A majority of such works either train the quantized network from scratch \cite{zhu2016trained,zhang2018lq,li2016ternary,hubara2016binarized,rastegari2016xnor,sakr2018true} or fine-tune a pre-trained model with quantization-in-the-loop \cite{jacob2018quantization,louizos2018relaxed,wang2019haq,yang2019quantization,baskin2018uniq,zhou2018explicit}. Where retraining is not an option, \cite{sakr2017analytical} provides analytical guarantees on the minimum precision requirements of a pre-trained FP network given a budget on the accuracy drop from FP. Training based quantization works fall into two classes of methods: 1) estimation based methods \cite{zhang2018lq,lin2017towards,li2016ternary,wang2019haq,jacob2018quantization}, where the full-precision weights and activations are quantized in the forward path, and gradients are back-propagated through a non-differentiable quantizer function via a gradient estimator such as the Straight Through Estimator (STE) \cite{bengio2013estimating}; and 2) optimization based methods, where gradients flow directly from the full-precision weights to the cost function via an approximate differentiable quantizer \cite{yang2019quantization,louizos2018relaxed,sakr2018true}, or by including an explicit quantization error term to the loss function \cite{hou2018loss,zhou2018explicit}. Application of these methods can be categorized into three clusters:

\textbf{Aggressive Quantization}: Methods such as binarization and ternarization have been highly successful for reducing DNN complexity. BinaryNets \cite{hubara2016binarized} quantize both weights and activations of DNNs to $\pm 1$, while XNORNets \cite{rastegari2016xnor} use a full-precision scalar to represent binarized weights in order to improve accuracy. Ternary Weight Networks (TWN) \cite{li2016ternary} quantize weights to $\{-1,0,1\}$ and leverage the resulting weight sparsity due to the '$0$' state to skip operations. Trained Ternary Quantization (TTQ) \cite{zhu2016trained} proposes learning the ternary scales via back-prop. However, a major drawback of such methods is the resulting accuracy loss especially when applied to lightweight DNNs such as MobileNet.
In Section \ref{sec:imagenet}, we show that ternarizing only the pointwise layers in MobileNetV1 on Imagenet, which correspond to $\sim 94\%$ of the total multiplication/additions (Table~\ref{tab:mobilenet-stats}), incurs a massive accuracy loss ($\sim 5.67 \% $) compared to the full-precision baseline.
Hence, such methods are typically benchmarked on simple datasets such as CIFAR-10, or use over parameterized models such as AlexNet \cite{krizhevsky2012imagenet} or ResNet-18 \cite{he2016deep} on ImageNet. In contrast, our proposed DBQ method is able to aggressively quantize the lightweight MobileNetV1 architecture with minimal loss in Top-1 accuracy (Fig~\ref{fig:motivation}).

\textbf{Non-uniform Quantization}: These methods seek to improve the performance of binarized/ternarized models while leveraging their arithmetic simplicity, e.g., LQNets \cite{zhang2018lq} and ABCNets \cite{lin2017towards}, by quantizing weights and activations as linear combinations of binary values. The resulting non-uniform multi-bit quantization allows the computation of dot products to be carried out using binary arithmetic with appropriate scaling and addition. However, these methods suffer from two major drawbacks: 1) the design of their quantization functions is computationally expensive as it requires an iterative solution of a non-convex optimization problem per-layer per-forward pass during training, which results in a significant training time overhead in the range $1.4\times$ $-$ $3.7\times$ \cite{zhang2018lq}; and 2) they suffer from gradient mismatch problems as they depend on the STE \cite{bengio2013estimating} method to compute the gradients during training. This renders the quantizer constructed by these methods to be sub-optimal, since they \textit{estimate} the quantizer parameters by minimizing a \emph{local} cost function, e.g., MSE. Moreover, these methods have been benchmarked only on over parameterized networks on ImageNet. Whereas our proposed DBQ method \emph{learns} the multiple \emph{ternary} branches by minimizing a \emph{global} loss function since the proposed quantizer is fully differentiable, which enables the efficient training of similar non-uniform quantizers, while also eradicating the need for any gradient estimator.

\textbf{Quantization of Lightweight DNNs}: Recent works that quantize MobileNets either apply fixed-point quantization with uniform \cite{louizos2018relaxed,sheng2018quantization,jacob2018quantization} or mixed \cite{wang2019haq,Uhlich2020Mixed} precision across layers. Hardware-Aware Quantization (HAQ) \cite{wang2019haq} proposes using reinforcement learning to learn the per-layer bit-precision for both weights and activations, whereas \cite{Uhlich2020Mixed} learns the bit-precision via a reformulation of the quantizer function and relying on the STE for gradient computation. Integer-Arithmetic-Only (IAO) \cite{jacob2018quantization} proposes using $8$-b quantization for accelerating the inference of MobileNets on hardware platforms such as Qualcom Hexagon and ARM NEON.  Relaxed Quantization (RQ) \cite{louizos2018relaxed} approximates the quantization function with a smooth differentiable approximate function, but the quantized values are still in fixed-point. Uniform Noise Injection Quantization (UNIQ) \cite{baskin2018uniq} proposes training a non-uniform quantizer using a special noise injection method that allows natural computation of gradients for quantized parameters. UNIQ uses a non-uniform quantizer requiring inefficient lookup tables and full precision multipliers/additions. Furthermore, all of these approaches use relatively high bitwidths ($\sim $ 6b$-$8b), and most even fail to bridge the accuracy gap between the quantized models and their full-precision baseline. In contrast, the proposed DBQ method is able to aggressively reduce the precision of the dominant ($94\%$) PW layers of MobileNetV1 to two ternary parameters with negligible degradation in the Top-1 accuracy.

\section{Differentiable Branched Quantizer (DBQ)}
\begin{figure}[t]
    \begin{center}
    \subfloat[]{\includegraphics[width=0.45\columnwidth]{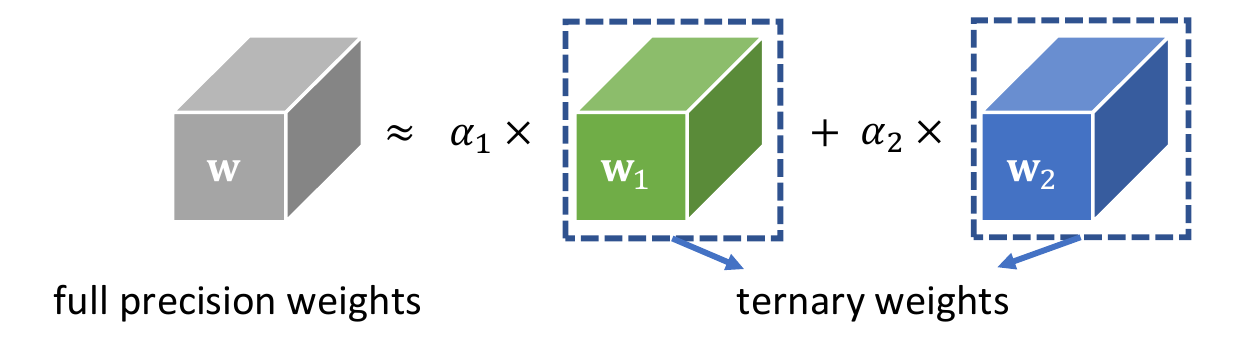}\label{fig:2t-explained}}%
    \qquad%
    \subfloat[]{\includegraphics[width=0.45\columnwidth]{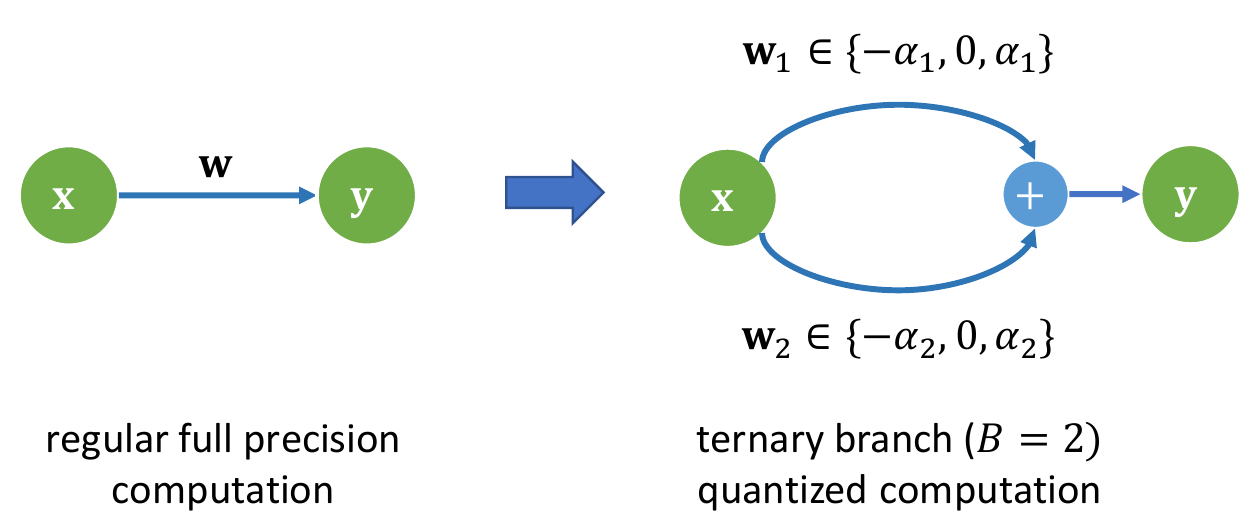}\label{fig:2t-branches}}%
    \end{center}
    \caption{Branched quantization of full precision weights: (a) as a linear combination of ternary weights, and (b) implemented as multiple parallel ternary branch operations to leverage the properties of ternary arithmetic for dot product computations.}%
    \label{fig:2t}%
\end{figure}

A ternary $B$-branch quantizer $Q(\mathbf{w})$ of a full precision weight vector $\mathbf{w} \in \reals^{D}$ (Fig~\ref{fig:2t}(a)) is given by:
\begin{equation}
    \mathbf{w}_q = Q(\mathbf{w})=\sum_{j=1}^{B}\alpha_j \mathbf{w}_j
    \label{eq:quant-eq}
\end{equation}
where $\mathbf{w}_j\in \{-1,0,1\}^D$ are the ternary branch weight vectors, and $\forall j \in [B]$: $\alpha_j >0$ are per-branch scalars. 
In DBQ, we wish to learn all the network parameters which requires the quantizer function $Q(\mathbf{w})$ to be made differentiable. To do so, we first formulate a parametric form of $Q(\mathbf{w})$ in Section~\ref{subsec:formulation} and then employ a smooth 'temperature-controlled' approximation of the quantizer step function to establish its differentiability in Section~\ref{subsec:differential}.

\subsection{Formulation of DBQ}\label{subsec:formulation}
We formulate the ternary $B$-branch quantizer in Fig.~\ref{fig:2t} as a $N=3^B$-level non-uniform quantizer $Q(\mathbf{w}): \reals^{D} \rightarrow \mathcal{V}^{D}$ with quantization levels $\mathcal{V}=\{v_i\}^{N}_{i=1}$. Assuming that the quantization levels $v_i$'s are sorted in ascending order, the $Q(\mathbf{w})$ can be written as a linear combination of $N-1$ step functions as shown below:
\begin{equation}\label{eq:proposed}
    Q(\mathbf{w}) = \sum_{i=1}^{N-1}\Big[\big(v_{i+1}-v_i\big) f(\mathbf{w}-t_i)\Big]-\frac{v_N-v_1}{2}
\end{equation}
where $f(\mathbf{u})=[\identityf{u_1>0},..., \identityf{u_{D}>0}]^{\text{T}}$ is an element-wise ideal step function, and $\{t_i\}_{i=1}^{N-1}$ are the quantizer thresholds. The $(v_N-v_1)/2$ term is the quantizer offset. We impose the ternary quantizer structure in \eqref{eq:quant-eq} via the constraint:
\begin{equation}\label{eq:constraint}
    v_i =\sum_{j=1}^Be_{i,j}\alpha_j
\end{equation}
where $e_{i,j} \in \{-1,0,1\}$, and thereby obtain the 
final quantizer expression:
\begin{equation}\label{eq:proposed2}
    Q(\mathbf{w}) = \gamma_2 \Bigg[\sum_{i=1}^{N-1}\Big[f(\gamma_1 \mathbf{w}-t_i)\sum_{j=1}^Bb_{i,j}\alpha_j\Big]-\sum_{j=1}^B\alpha_j\Bigg]
\end{equation}
where $b_{i,j}=e_{i+1,j}-e_{i,j} \in \{-2,-1,0,1,2\}$ $\forall j \in [B]$ are \textit{fixed} coefficients, and $\gamma_1\ \&\ \gamma_2$ are pre/post-quantization scales to ensure that the quantizer operates on normalized inputs. Thus, the branched quantizer is parametrized by $\mathcal{P}_Q=\{\alpha_1, ..., \alpha_B, \gamma_1, \gamma_2, t_1, ..., t_{N-1}\}$ and these all need to be learned.

\begin{figure}[t]
    \begin{center}
    \includegraphics[width=0.5\linewidth]{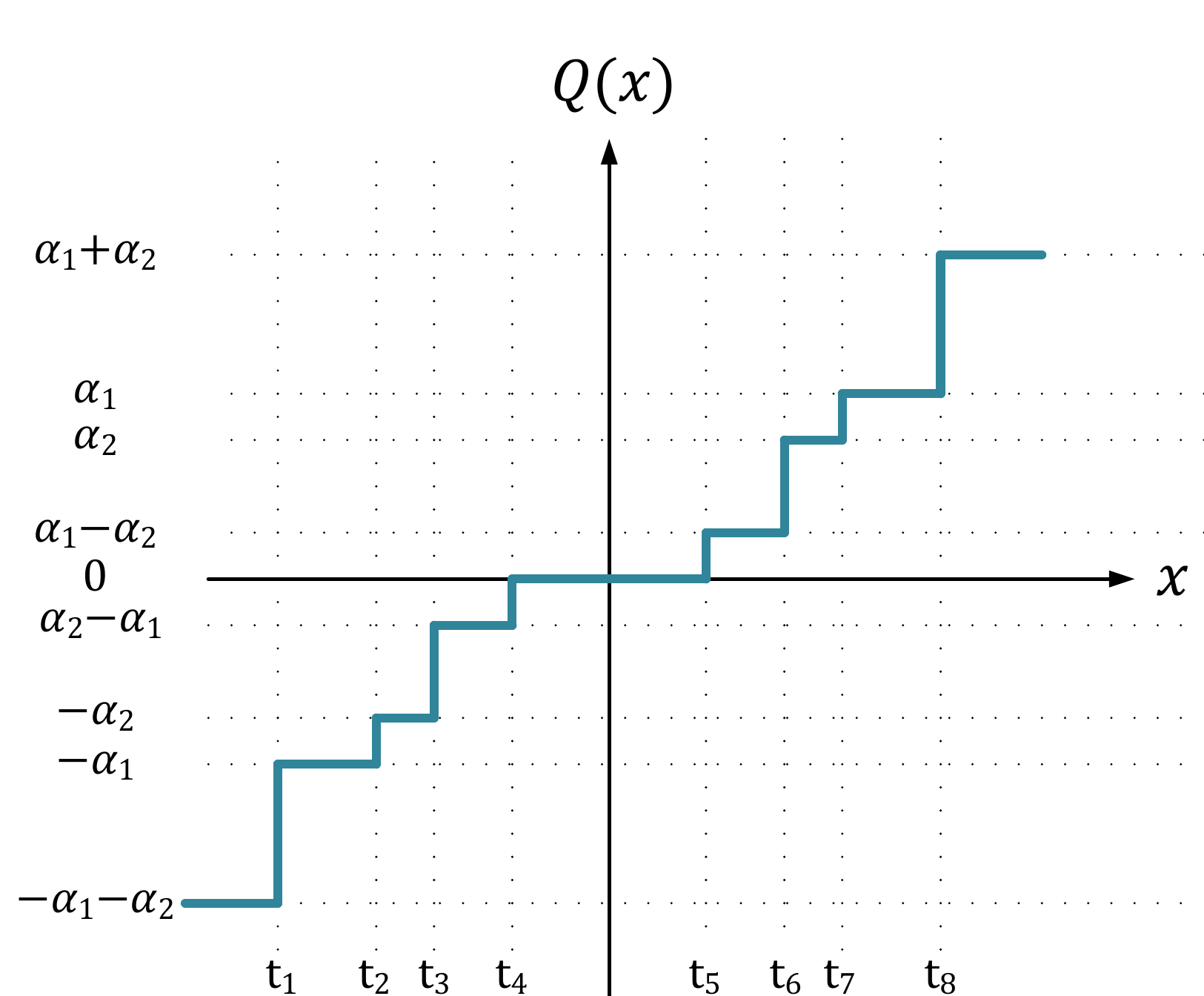}%
    \end{center}
    \caption{Visualization of a two ternary (2T) branch quantizer with branch scales $\alpha_1$ and $\alpha_2$ assuming $\alpha_1\geq \alpha_2 \geq \frac{\alpha_1}{2} \geq 0$.}%
    \label{fig:2t-quant}%
\end{figure}

In this paper, we focus on the $B=2$ case, i.e., two ternary branch, as visualized in Fig.~\ref{fig:2t-quant}, with $N=3^2=9$ different quantization levels $v_i$. In this case, \eqref{eq:proposed2} can be expanded as:
\begin{align}
    \begin{split}
        Q(\mathbf{w}) &= \gamma_2 \Big[\alpha_2 f(\gamma_1 \mathbf{w} -t_1) + (\alpha_1 - \alpha_2) f(\gamma_1 \mathbf{w} -t_2) + (2\alpha_2 - \alpha_1) f(\gamma_1 \mathbf{w} -t_3) \\ &+  (\alpha_1 - \alpha_2) f(\gamma_1 \mathbf{w} -t_4) + (\alpha_1 - \alpha_2) f(\gamma_1 \mathbf{w} -t_5)  + (2\alpha_2 - \alpha_1) f(\gamma_1 \mathbf{w} -t_6) \\
        &+ (\alpha_1 - \alpha_2) f(\gamma_1 \mathbf{w} -t_7)  + \alpha_2 f(\gamma_1 \mathbf{w} -t_8) -(\alpha_1 + \alpha_2)\Big]
    \end{split}
\end{align}

\subsection{Differentiability}\label{subsec:differential}
Inspired by \cite{yang2019quantization,xie2018snas}, we replace the non-differentiable $f$ in \eqref{eq:proposed2} with a smooth sigmoid approximation $\hat{f}_T$ as follows:
\begin{align}
\begin{split}
   \hat{f}_T(u) &= \frac{1}{1+\text{exp}(-Tu)} \\
\end{split}
\end{align}
where the \emph{temperature} parameter $T$ controls the approximation error, i.e.,:
\begin{equation}\label{eq:error}
    e_T(u) = \hat{f}_T(u) - f(u) \xrightarrow[T \to \infty]{} 0
\end{equation}
When learning the quantizer parameters $\mathcal{P}_Q$, the temperature $T$ is increased gradually as the training converges so that $\hat{f}_T(u)\rightarrow f(u)$. The resultant differentiable quantizer $Q_T(\mathbf{w})=\mathbf{w}_q=\mathbf{z}$ therefore enables a straightforward calculation of the gradients for all quantizer and model parameters w.r.t. loss function $\mathcal{L}$ as follows:

\begin{align}
    \frac{\partial \calL}{\partial \gamma_2} &= \frac{1}{\gamma_2} \sum_{k=1}^D\frac{\partial \calL}{\partial z_k}z_k \label{eq:deriv1}\\
    \frac{\partial \calL}{\partial \alpha_j} &= \gamma_2 \sum_{k=1}^D \frac{\partial \calL}{\partial z_k} \Bigg[\sum_{i=1}^{N-1}\Big[b_{i,j}g_{k,i}\Big]-1\Bigg] \label{eq:deriv2}\\
    \frac{\partial \calL}{\partial t_i} &= -\gamma_2 T \sum_{k=1}^D \frac{\partial \calL}{\partial z_k} \Big[h_{k,i}\sum_{j=1}^Bb_{i,j}\alpha_j \Big] \label{eq:deriv3}\\
     \frac{\partial \calL}{\partial w_k} &= \gamma_1 \gamma_2 T  \frac{\partial \calL}{\partial z_k} \sum_{i=1}^{N-1} \Big[ h_{k,i}\sum_{j=1}^Bb_{i,j}\alpha_j \Big] \label{eq:deriv4}\\  
    \frac{\partial \calL}{\partial \gamma_1} &= \gamma_2 T \sum_{k=1}^D \frac{\partial \calL}{\partial z_k}w_k \Bigg[ \sum_{i=1}^{N-1} \Big[ h_{k,i}\sum_{j=1}^Bb_{i,j}\alpha_j \Big] \Bigg] \label{eq:deriv5}
\end{align}
where $h_{k,i} = g_{k,i}(1-g_{k,i})$ and $g_{k,i} = \hat{f}_T(\gamma_1w_k-t_i)$ for brevity. By doing so, we eliminate the need for the STE and the expensive computational overhead introduced in estimation-based methods such as LQNet \cite{zhang2018lq} or ABCNet \cite{lin2017towards}. Note that software frameworks such as PyTorch \cite{paszke2017automatic} automatically take care of computing these gradients so these don't need to be explicitly coded.

\subsection{Implementation Details}
\textbf{Parameter Initialization}: Initializing the quantizer parameters $\mathcal{P}_Q$ is performed once before training and requires an initial vector $\mathbf{w}\in \reals^D$ which can be from a pre-trained network or from random initialization (training from scratch). The initialization procedure is as follows: 1) the post-quantization scale $\gamma_2$ is set to the maximum absolute value in $\mathbf{w}$, and the pre-quantization scale $\gamma_1$ is set to $1/\gamma_2$. This ensures that the quantizer operates on normalized parameters which facilitates the optimization of its parameters, and that the quantized values are of the same scale as the inputs; 2) to find the optimal thresholds $\{t_i\}_{i=1}^{N-1}$, we first compute the optimal $N$ centroids $\{c_i\}_{i=1}^{N}$ of the normalized vector $\gamma_1\mathbf{w}$ via $k$-means, and then $\forall i\in[N-1]$ we set $t_i$ to be the midpoint of the interval $[c_i, c_{i+1}]$; and 3) a good initialization for $\{\alpha_j\}_{j=1}^{B}$ is found by solving for the optimal values that minimize the $L_2$ norm between the normalized vector $\gamma_1\mathbf{w}$ and its quantized counterpart.

\textbf{Training and Inference}: During training, the proposed DBQ quantizer is used with the approximate smooth step function $\hat{f}_T$ for both forward and backward calculations (\eqref{eq:proposed2} \& \eqref{eq:deriv1}$-$\eqref{eq:deriv5}). For a given layer in the network that performs the function $\mathbf{y}=F(\mathbf{w}, \mathbf{x})$, applying DBQ simply boils down to composing the quantizer described in \eqref{eq:proposed2} with the function $F$: $\mathbf{y}=F(Q_T(\mathbf{w}), \mathbf{x})$. For quantizing convolutional layers, we apply kernel-wise quantizers. The overhead of full precision scales is amortized across the large filter lengths. The choice of the temperature parameter $T$ is important. A large value of $T$ would reduce the approximation error in \eqref{eq:error}, however the gradients would saturate quickly, thus causing a bottleneck for learning the quantizer parameters. Therefore, an initial small value for $T$ is used for the first training epoch, and its value is increased for successive epochs based on a pre-determined temperature update schedule. A simple yet effective schedule is to linearly increment the temperature with the number of epochs: $T = T_{\text{init}} + e\times T_{\text{inc}}$. During inference, the approximate step function is replaced with the ideal function $f$ such that the quantizer output satisfies \eqref{eq:quant-eq}.

\begin{figure}[t]
    \begin{center}
    \includegraphics[width=0.99\linewidth]{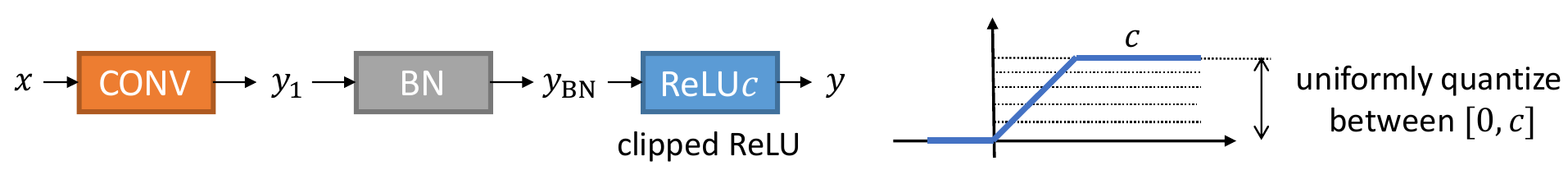}
    \end{center}
    \caption{Quantizing activations post-ReLU requires a pre-determined clipping parameter $c$.}%
    \label{fig:act-quant}%
\end{figure}

\textbf{Activation Quantization}:
The challenge in quantizing input activations with a fixed-point quantizer during training  is determining a suitable clipping value (Fig.~\ref{fig:act-quant}). Traditionally, the use of ReLU$6$ (which clips at $6$) has been a popular choice due to its simplicity \cite{sheng2018quantization,jacob2018quantization}. However, the choice of $6$ provides no guarantees on the clipping probability, and can therefore yield sub-optimal results. Similar to \cite{dbouk2020low}, we propose clipping the post-BN activations $y_{\text{BN}}$ (Fig.~\ref{fig:act-quant}) using:
\begin{equation}
    c = \max_{i\in[C]}(\beta_l^{(i)}+k\gamma_l^{(i)})
    \label{eq:c}
\end{equation}
where $C$ is the number of channels in the activation tensor $y_1$, $(\beta^{(i)},\gamma^{(i)})$ are learnable per-channel shift and scale parameters of BN, and $k$ is a network hyperparameter that controls the clipping probability. Assuming that the distribution of $y_{\text{BN}}^{(i)} \sim \mathcal N\big(\beta^{(i)},(\gamma^{(i)})^2\big)$  \cite{ioffe2015batch} and using $6\sigma$ rule ($k=6$), one can show that the choice of $c$ in \eqref{eq:c} guarantees:
\begin{equation}
    \text{Pr}\{y_{\text{BN}}\leq c\} \geq 0.999
\end{equation}
Note that having a fixed clipping value $c$ for all channels is crucial in order to ensure that the dot product operations can be implemented in fixed-point.

\section{Experimental Results}
To demonstrate the effectiveness of the DBQ method for quantizing lightweight networks, we evaluate it on three different image classification datasets: 1) CIFAR-10 \cite{krizhevsky2009learning} using ResNet-20 \cite{he2016deep}; 2) ImageNet (ILSVRC 2012) \cite{russakovsky2015imagenet} using MobileNetV1 \cite{howard2017mobilenets}, MobileNetV2 \cite{sandler2018mobilenetv2}, and ShuffleNetV2 \cite{ma2018shufflenet}; and 3) the recently proposed Visual Wake Words \cite{chowdhery2019visual} using MobileNetV1. In all of our experiments, we train full precision models from scratch, and perform fine tuning on said models for training their quantized counterparts. We use stochastic gradient descent for training all the models. For further details on the training setup for each experiment, please check the supplementary material.

\subsection{Complexity Metrics}
We propose a set of metrics, inspired by those used in \cite{sakr2017analytical,sakr2018per}, in order to quantify the complexity reduction achieved by our proposed method.

\textbf{Computational Cost} {($\bm{\mathcal{C}_C}$)} for an $L$-layer network:
\begin{align}
\begin{split}\label{eq:cc}
     \calC_C &= \sum_{l=1}^LN_l\Big[D_lB_{W,l}B_{A,l}  + (D_l-1)(B_{A,l}+B_{W,l}+\lceil\log_2D_l \rceil-1)\Big]   
\end{split}
\end{align}
where $N_l$ is the number of $D_l$-dimensional dot products in layer $l$ with $B_{W,l}$ and $B_{A,l}$ being the weights and activations precisions respectively. This cost essentially measures the number of 1b full adders (FAs) needed to implement the dot products required for a given network. For full precision (32b) parameters, we make the simplifying assumption of treating them as 23b (mantissa precision) fixed-point parameters.

\textbf{Sparsity-Aware Computational Cost} {($\bm{\mathcal{C}_S}$)} is also defined in order to leverage weight-sparsity in different models that can be reflected on the model complexity:
\begin{align}
\begin{split}\label{eq:sc}
     \calC_S &= \sum_{l=1}^LN_l\Big[D'_lB_{W,l}B_{A,l}  + (D'_l-1)(B_{A,l}+B_{W,l}+\lceil\log_2D_l \rceil-1)\Big]   
\end{split}
\end{align}
where $D'_l$ is the number of non-zero weights in the corresponding $D_l$-dimensional dot product.

\textbf{Representational Cost} {($\bm{\mathcal{C}_R}$)} for an $L$-layer network:
\begin{align}
\begin{split} \label{eq:rc}
     \calC_R &= \sum_{l=1}^L\Big[|W_l|B_{W,l} + |A_l|B_{A,l}\Big] 
\end{split}
\end{align}
where $|W_l|$ and $|A_l|$ are the number of elements in the weight and activation tensors in layer $l$, respectively.

\textbf{Model Storage Cost} {($\bm{\mathcal{C}_M}$)} for an $L$-layer network:
\begin{align}
\begin{split} \label{eq:mc}
     \calC_M &= \sum_{l=1}^L|W_l|B_{W,l} 
\end{split}
\end{align}

which only accounts for the weight storage, and can be useful for studying model compression.

\begin{table}[t]
    \begin{center}
    \resizebox{0.6\columnwidth}{!}{%
    \renewcommand{\arraystretch}{1.2}
    \begin{tabular}{l l c c}
    \clineB{1-4}{2.5}
    \textbf{Method} & \textbf{Acc}. ($\bm{\Delta}$) [\%] & $\bm{\mathcal{C}_C}\ (\bm{\mathcal{C}_S})$ [$10^{9}$FA] & $\bm{\mathcal{C}_R}\ (\bm{\mathcal{C}_M})$ [$10^{6}$b]\\\clineB{1-4}{2.5}
    FP \cite{zhang2018lq} & $92.10\ (/)$ & $23.73\ (23.73)$ & $14.63\ (8.63)$\\\hline
    LQNet-1B \cite{zhang2018lq} & $90.10\ (-2.171)$& $1.60\ (1.60)$ & $6.34\ (0.35)$
    \\\hline
    LQNet-2B \cite{zhang2018lq} & $91.80\ (-0.325)$& $2.83\ (2.83)$ & $6.61\ (0.61)$
    \\\hline
    LQNet-3B \cite{zhang2018lq} & $92.00\ (-0.108)$& $4.07\ (4.07)$ & $6.88\ (0.88)$
    \\\hline\hline
    FP (Ours) & $92.00\ (/)$ & $23.73\ (23.73)$ & $14.63\ (8.63)$\\\hline
    DBQ-1T (Ours) & $\mathbf{91.06\ (-1.021)}$ & $\mathbf{1.60\ (0.92)}$ & $6.61\ (0.61)$\\\hline
    DBQ-2T (Ours) & $\mathbf{91.93\ (-0.076)}$ & $\mathbf{2.83\ (1.79)}$ & $7.15\ (1.15)$\\\hline
    \end{tabular}
    }
    \end{center}
    \caption{The accuracy on CIFAR-10 and complexity metrics ($\mathcal{C}_C$, $\mathcal{C}_S$, $\mathcal{C}_R$, $\mathcal{C}_M$) for ResNet-20 using our method DBQ compared to LQNet. $\Delta$ represents the normalized accuracy drop of the quantized models with respect to its full precision baseline. The first, last layers, and input activations are kept in full precision for the quantized models in accordance with \cite{zhang2018lq}.}
    \label{tab:resnet20}
\end{table}

\subsection{CIFAR-10 Results}
We first demonstrate the effectiveness of DBQ on the CIFAR-10 dataset using the popular network ResNet-20 \cite{he2016deep}. To ensure a fair comparison with the LQNet \cite{zhang2018lq} models, we do not quantize the first and last fully connected layers, and we keep all activations in full precision. Table~\ref{tab:resnet20} summarizes the accuracy (and percentage drop) as well as the four complexity metrics ($\mathcal{C}_C$, $\mathcal{C}_S$, $\mathcal{C}_R$, $\mathcal{C}_M$) for different number of branches used. At iso-number of branches, the DBQ models achieve higher accuracies for the same $\mathcal{C}_{C}$ and lower $\mathcal{C}_{S}$ due to the high number of zero valued weights, as opposed to binary branches where the weights are either $\pm 1$. Comparing the DBQ-2T and LQNet-3B models, which achieve comparable accuracies, DBQ-2T requires $\sim 32\%$ less $\calC_{C}$ and $\sim 56 \%$ less $\calC_{S}$, at the expense of an extra bit per-parameter, which is reflected in the marginal $\sim 4 \%$ increase in $\calC_R$.

\subsection{ImageNet Results}
\label{sec:imagenet}
In this section, we report results for MobileNetV1 \cite{howard2017mobilenets}, MobileNetV2 \cite{sandler2018mobilenetv2}, and ShuffleNetV2 \cite{ma2018shufflenet} on ImageNet. We first focus on MobileNetV1 by performing an ablation study, and leverage these results for quantizing the more recent MobileNetV2 and ShuffleNetV2.

\begin{table}[t]
    \begin{center}
    \resizebox{\columnwidth}{!}{%
    \renewcommand{\arraystretch}{1.2}
    \begin{tabular}{l|c c c c c|c c c}
    \clineB{1-9}{2.5}
    \textbf{Model Name} & \textbf{Activations}&\textbf{FL} & \textbf{DW} & \textbf{PW} & \textbf{FC} &  \textbf{Top-1/5 Acc.} [$\%$] & $\bm{\mathcal{C}_C}\ (\bm{\mathcal{C}_S})$ [$10^{10}$FA] & $\bm{\mathcal{C}_R}\ (\bm{\mathcal{C}_M})$ [$10^{7}$b]  \\ \clineB{1-9}{2.5}
    FP & ReLU - 32b     & 32b & 32b & 32b & 32b & $\mathbf{72.12/90.43}$ &$33.37\ (33.37)$ & $30.00\ (13.54)$\\ \hline
    FX8-1 & ReLU6 - 8b     & 32b & 8b & 8b & 32b & $71.65/90.17$  & $5.78\ (5.39)$ & $10.38\ (5.90)$\\ \hline
    FX8-2 & ReLU6 - 8b     & 8b & 8b & 8b & 8b & $71.60/90.19$  & $5.24\ (4.85)$ & $7.56\ (3.44)$\\ \hline
    FX8-3 & ReLU$x$ - 8b     & 8b & 8b & 8b & 8b & $\mathbf{71.86/90.26}$  & $5.24\ (4.85)$&  $7.56\ (3.44)$ \\ \hline
    DBQ-1T & ReLU - 32b     & 32b & 32b & 1T & 32b & $66.45/86.72$   & $3.60\ (2.61)$&  $20.58\ (4.12)$\\ \hline
    DBQ-2T-1 & ReLU - 32b     & 32b & 32b & 2T & 32b & $71.09/89.71$   & $5.23\ (3.77)$ & $21.21\ (4.75)$\\ \hline
    DBQ-2T-2 & ReLU6 - 8b     & 32b & 8b & 2T & 32b & $70.25/89.42$   &$2.73\ (1.97)$ & $9.12\ (4.64)$\\ \hline
    DBQ-2T-3 & ReLU$x$ - 8b     & 32b & 8b & 2T & 32b & $70.80/89.75$  &$2.73\ (1.97)$ & $9.12\ (4.64)$\\ \hline
    DBQ-2T-4 & ReLU$x$ - 8b    & 8b &8b & 2T & 8b & $\mathbf{70.92/89.61}$  & $\mathbf{2.18\ (1.42)}$ & $\mathbf{6.30\ (2.18)}$\\\hline 
    \end{tabular}
    }
    \end{center}
    \caption{The Top-1/5 accuracy on ImageNet and complexity metrics ($\mathcal{C}_C$, $\mathcal{C}_S$, $\mathcal{C}_R$, $\mathcal{C}_M$) for MobileNetV1 under different precision assignments. Models denoted by DBQ-$z$T are trained using our differentiable branch quantizer with $B=z$ ternary branches. ReLU$x$ denotes a clipped ReLU using our proposed clipping method in Eq.~\eqref{eq:c}.}
    \label{tab:imagenet-results}
\end{table}
\textbf{Ablation Study:} Table~\ref{tab:imagenet-results} summarizes the Top-1,5 accuracies of all the MobileNetV1 models trained with different layer precision assignments in order to evaluate the impact of our design choices. To see the impact of using two ternary branches instead of one, we begin with the DBQ-1T model which is obtained by quantizing only the PW layers of MobileNetV1 to one ternary branch (1T) keeping all other activations and weights in full precision. Table~\ref{tab:imagenet-results} shows that DBQ-1T achieves a massive $89\%$ reduction in $\calC_C$ compared to the FP model but at a catastrophic loss of $5.67\%$ in the Top-1 accuracy. In contrast, DBQ-2T-1, which is DBQ-1T with a second ternary branch, is able to recover accuracy to within $1.03\%$ of the full-precision baseline while also achieving massive savings in $\mathcal{C}_C$ of $84 \%$. Quantizing the activations and the remaining layers weights of DBQ-2T-1 to 8b fixed-point, i.e., DBQ-2T-4, incurs a minimal loss in accuracy of $1.2\%$ compared to the FP model while also achieving even greater reduction in both $\mathcal{C}_C$ ($93\%$) and $\mathcal{C}_R$ ($70\%$). The reduction in $\mathcal{C}_S$ increases to $96 \%$ when branch sparsity is exploited to skip computations. 

Note that the reason that only PW layers are quantized using ternary branches is three-fold: 1) PW layers consume $\sim 94\%$ of the amount of multiply-adds required for inference (Table~\ref{tab:mobilenet-stats}); 2) we have observed that quantizing the PW layers has the most severe impact on classification accuracy compared to quantizing other layers; and 3) DW layers suffer from extremely small dot-product lengths (9), rendering them unsuitable for multiple branch quantization (the overhead of branch-merge and scaling operations will dominate).

The benefits of our proposed BN-based clipping described in \eqref{eq:c} can be seen by comparing the accuracy of the 8b fixed-point model FX8-3 using BN-based clipping with $k=6$ with its ReLU6-based clipping counterpart FX8-2. The Top-1 accuracy of FX8-3 is better than FX8-2 without any overhead in training or inference. Similarly for DBQ-2T-3 and DBQ-2T-2.

\begin{table}[t]
    \begin{center}
    \resizebox{\columnwidth}{!}{%
    \renewcommand{\arraystretch}{1.2}
    \begin{tabular}{l|c c c c c|c c c}
    \clineB{1-9}{2.5}
    \textbf{Method} & \textbf{Act.} &   \textbf{FL} & \textbf{DW} & \textbf{PW}  & \textbf{FC} &  \textbf{Top-1 Acc.} [$\%$] & $\bm{\mathcal{C}_C}\ (\bm{\mathcal{C}_S})$ [$10^{10}$FA] & $\bm{\mathcal{C}_R}\ (\bm{\mathcal{C}_M})$ [$10^7$b]\\ \clineB{1-9}{2.5}
    IAO$^\star$ \cite{jacob2018quantization} & 8b & 8b & 8b & 8b & 8b & $\mathbf{69.00}^*$ & $4.97\ (/)$ & $7.49\ (3.37)$
    \\ \hline
    UNIQ \cite{baskin2018uniq} & 8b & 5b & 5b & 5b & 5b & $67.50$ & $3.70\ (/)$ & $6.29\ (2.18)$
    \\ \hline
    UNIQ \cite{baskin2018uniq} & 8b & 4b & 4b & 4b & 4b & $66.00$ & $3.19\ (/)$ & $5.87\ (\mathbf{1.76})$
    \\ \hline
    UNIQ \cite{baskin2018uniq} & 8b & 8b & 8b & 8b & 8b & $68.25$ & $5.24\ (/)$ & $7.56\ (3.44)$
    \\ \hline
    QSM$^\star$ \cite{sheng2018quantization} & 8b & 8b & 8b & 8b & 8b & $68.03$ & $4.97\ (/)$ & $7.49\ (3.37)$
    \\ \hline
    RQ \cite{louizos2018relaxed} & 5b & 5b & 5b & 5b & 5b & $61.50$ & $\mathbf{2.68}\ (/)$ & $\mathbf{4.75}\ (2.18)$
    \\ \hline
    RQ \cite{louizos2018relaxed} & 6b & 6b & 6b & 6b & 6b & $67.50$ & $3.42\ (/)$ & $5.69\ (2.60)$
    \\ \hline
    HAQ cloud \cite{wang2019haq} & mixed & 8b & mixed & mixed & 8b & $65.33 - 71.20^{\dagger}$ & $2.73\ (/)$ & $5.09\ (3.12)$
    \\ \hline
    HAQ edge \cite{wang2019haq} & mixed & 8b & mixed & mixed & 8b & $67.40 - 71.20^\dagger$ & $4.06\ (/)$ & $5.87\ (2.49)$
    \\ \hline \hline
    FX8 (Ours)& 8b & 8b & 8b & 8b & 8b & $\mathbf{71.86}$& $5.24\ (4.85)$&  $7.56\ (3.44)$\\ \hline
    DBQ-2T (Ours)& 8b & 8b & 8b & 2T & 8b & $\mathbf{70.92}$ & $\mathbf{2.18\ (1.42)}$ & $6.30\ (2.18)$\\\hline 
    \multicolumn{9}{l}{$^{\star}$models with BN folding \qquad $^{*}$results extracted from a plot \qquad $^{\dagger}$exact accuracy not reported} 
    \end{tabular}
    }
    \end{center}
    \caption{The Top-1 accuracy on ImageNet and complexity metrics ($\mathcal{C}_C$, $\mathcal{C}_S$, $\mathcal{C}_R$, $\mathcal{C}_M$) for MobileNetV1 using our method (DBQ-2T) compared to state-of-the art training-based quantization methods.}
    \label{tab:imagenet-comparison}
\end{table}

\begin{wrapfigure}{R}{0.5\textwidth}
  \begin{center}
    \includegraphics[width=0.48\textwidth]{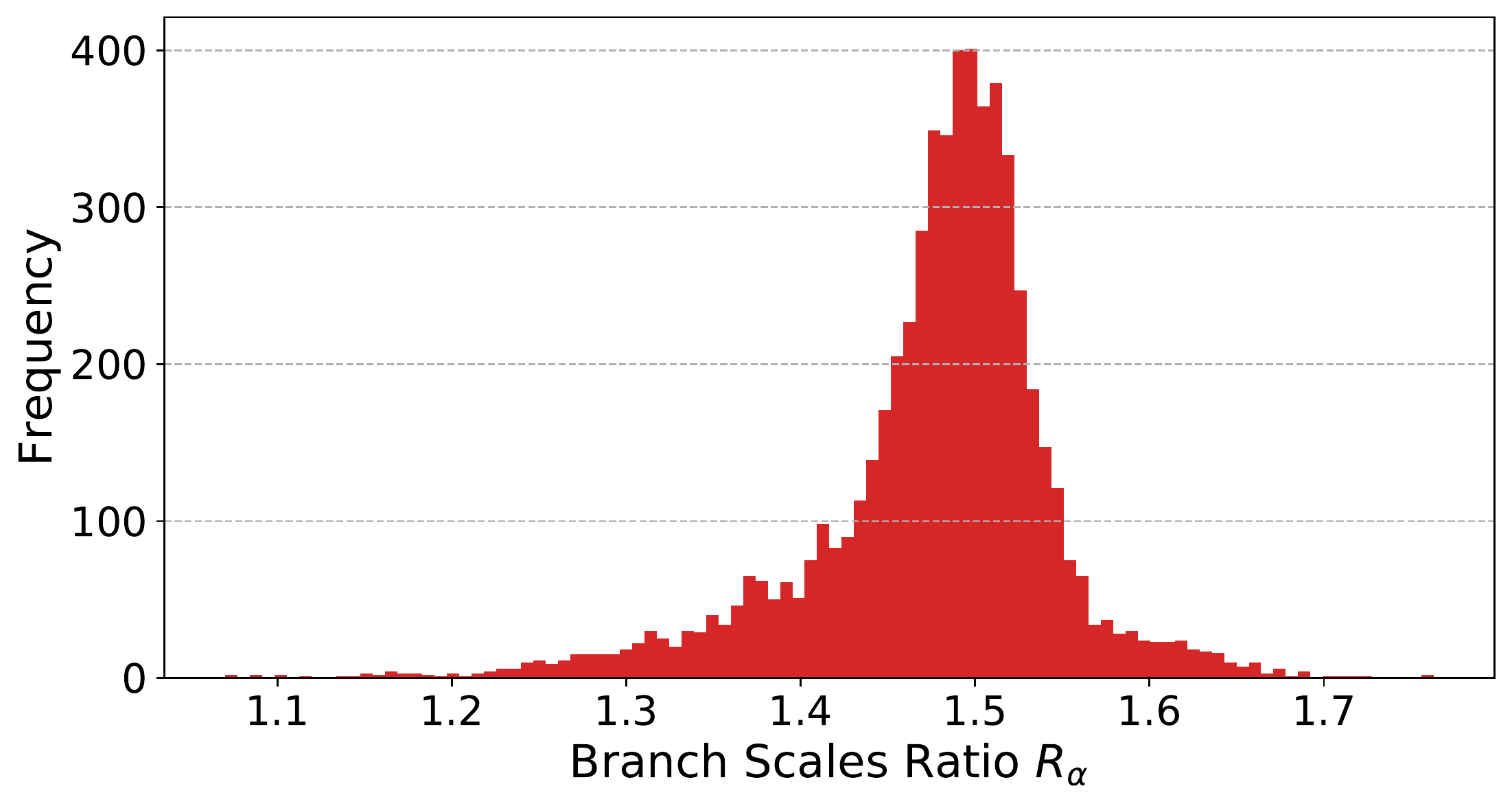}
  \end{center}
  \caption{The distribution of the ratio of the ternary branch scales $\alpha_1$ and $\alpha_2$ for DBQ-2T-4 from Table~\ref{tab:imagenet-results}.}
   \label{fig:hist-scales}
\end{wrapfigure}

\textbf{Branching Utility:} A 2T quantizer should result in $9$ distinct quantization levels as shown in Fig.~\ref{fig:2t-quant}. However, in a 2T branched quantizer such as ours, it is possible for the number of quantization levels to be smaller than 9, e.g., if $\alpha_1=\alpha_2$ then the number of quantization levels is 5. In this case, the full representational power of the 2T branched quantizer is not utilized. To see if the 2T branched quantizer generates all 9 levels, we plot the distribution of the ratio $R_{\alpha}=\frac{\alpha_1}{\alpha_2}$ across all the PW layers in the DBQ-2T-4 model (Table~\ref{tab:imagenet-results}). The distribution is centered around $R_{\alpha}=1.48$ with more than $99\%$ of the values lying in the range $[1.2,1.7]$. This demonstrates that the quantizer learned by DBQ employs the full representational power offered by the 2T structure.

\textbf{Comparison with State-of-the Art:} Table~\ref{tab:imagenet-comparison} compares the performance of our proposed DBQ method against state-of-the art results on ImageNet for MobileNetV1. Our model DBQ-2T, which corresponds to DBQ-2T-4 in Table~\ref{tab:imagenet-results} achieves the lowest computational cost $\mathcal{C}_C$ ($2.18 \times 10^{10}$ FAs) compared to previously published networks, while achieving the highest Top-1 accuracy $70.92\%$. Compared to the lowest complexity model RQ \cite{louizos2018relaxed}, DBQ-2T achieves a $19\%$ reduction in $\mathcal{C}_C$ with a $9.42\%$ improvement in Top-1 accuracy at iso-storage complexity $\mathcal{C}_M$. Furthermore, DBQ-2T improves upon the accuracy of the IAO model \cite{jacob2018quantization}, which currently achieves the highest Top-1 accuracy, by $1.92\%$ but with a massive reduction in complexity $\mathcal{C}_C$ ($56\%$), $\mathcal{C}_R$ ($16\%$), and $\mathcal{C}_M$ ($35\%$).

\textbf{More Lightweight Networks:} Table \ref{tab:extra-results} demonstrates the performance of DBQ when applied to the more recent lightweight networks: MobileNetV2 and ShuffleNetV2. Similar to MobileNetV1, we find that the PW layers \textit{dominate} the number of operations required for a single inference for both MobileNetV2 ($87\%$) and ShuffleNetV2 ($90\%$). Thus, and inline with our experiments on MobileNetV1, we quantize all PW layers using 2T, with the remaining layers and activations quantized to 8b fixed-point. We observe a minimal $1.3\%$ (MobileNetV2) and $2.6\%$ (ShuffleNetV2) drop in accuracy compared to FP, while achieving \textit{massive} ($77\% - 95\%$) reductions in all the complexity metrics. A comparison between DBQ and \cite{Uhlich2020Mixed} for MobileNetV2 is present in the supplementary material.

\begin{table}[t]
    \begin{center}
    \resizebox{\columnwidth}{!}{%
    \renewcommand{\arraystretch}{1.2}
    \begin{tabular}{l|c c c c c|c c c}
    \clineB{1-9}{2.5}
    \textbf{Model} & \textbf{Act.} &   \textbf{FL} & \textbf{DW} & \textbf{PW}  & \textbf{FC} &  \textbf{Top-1 Acc.} [$\%$] & $\bm{\mathcal{C}_C}\ (\bm{\mathcal{C}_S})$ [$10^{10}$FA] & $\bm{\mathcal{C}_R}\ (\bm{\mathcal{C}_M})$ [$10^7$b]\\ \clineB{1-9}{2.5}
    MobileNetV2-FP & 32b& 32b & 32b & 32b & 32b & $71.88$ & $17.83\ (17.83)$ & $32.87\ (11.22)$ \\ \hline
    MobileNetV2-2T & 8b & 8b & 8b & 2T & 8b& $\mathbf{70.54}$ & $\mathbf{1.42}\ (\mathbf{1.11})$ & $\mathbf{7.45}\ (\mathbf{2.04})$ \\ \hline \hline
    ShuffleNetV2-FP & 32b& 32b & 32b & 32b & 32b & $69.36$ & $8.52\ (8.52)$ & $13.81\ (7.29)$ \\ \hline
    ShuffleNetV2-2T & 8b & 8b & 8b & 2T & 8b& $\mathbf{ 66.74}$ & $\mathbf{0.64}\ (\mathbf{0.46})$ & $\mathbf{3.21}\ (\mathbf{1.38})$ \\ \hline 
    \end{tabular}
    }
    \end{center}
    \caption{The Top-1 accuracy on ImageNet and complexity metrics ($\mathcal{C}_C$, $\mathcal{C}_S$, $\mathcal{C}_R$, $\mathcal{C}_M$) for MobileNetV2 and ShuffleNetV2 using our method (DBQ-2T).}
    \label{tab:extra-results}
\end{table}

\subsection{Visual Wake Words Results}
We study the accuracy-precision-complexity trade-off in quantized DNNs using the Visual Wake Words (VWW) dataset that was recently proposed by Google \cite{chowdhery2019visual} in order to facilitate the development of lightweight vision models for deployment on resource-constrained Edge devices.
This dataset reflects a typical real-world scenario involving the detection of specific events by observing incoming data, e.g., monitoring a camera video feed in order to detect the presence of a person \cite{chowdhery2019visual}, similar to the use of audio wake words in speech recognition. The VWW dataset is derived from the COCO dataset \cite{lin2014microsoft} via a simple re-labeling of the available images has a training set of 115k images and a test set with 8k images.

\begin{figure}[t]
    \begin{center}
        \subfloat[]{\includegraphics[height=4.2cm]{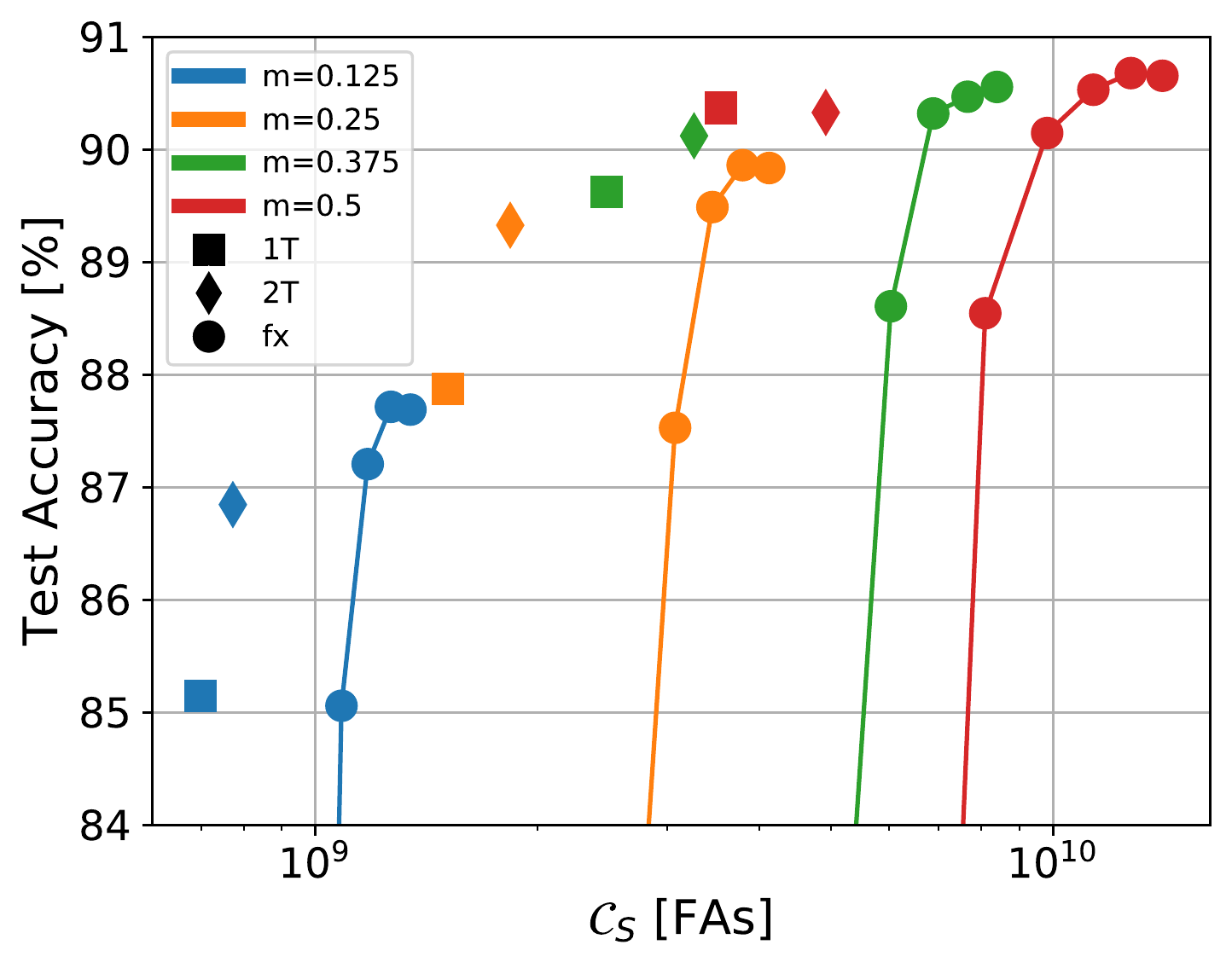}\label{fig:vww-cc}}%
        \qquad%
        \subfloat[]{\includegraphics[height=4.2cm]{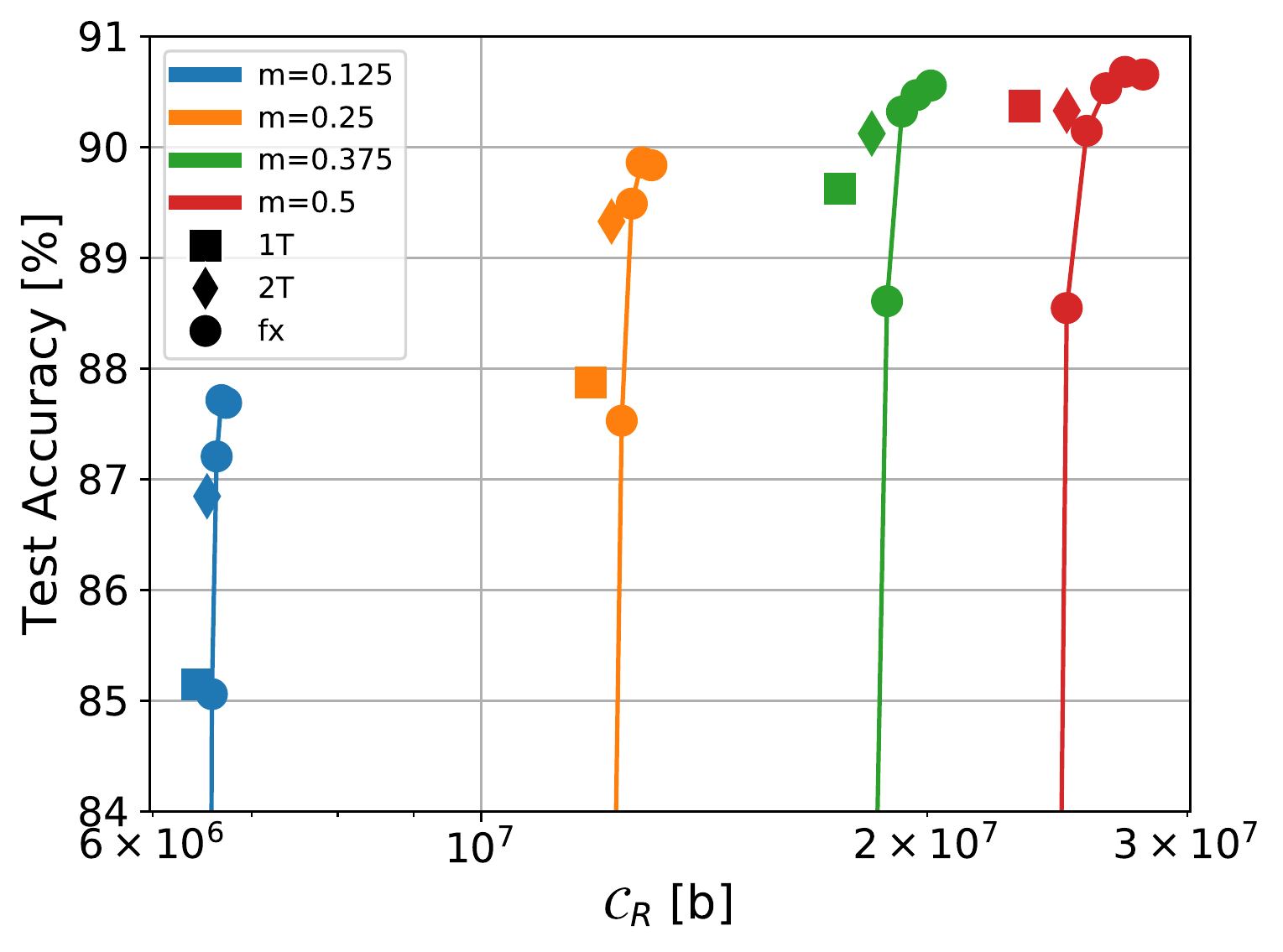}\label{fig:vww-rc}}%
    \end{center}
    
    \caption{The test accuracy of MobileNetV1 on the Visual Wake Words dataset with varying precision assignment and width multiplier $m$ vs. (a) sparsity-aware computational cost, and (b) representational cost. Only the precision of the pointwise layer's weights are changing, whereas all the remaining activations and weights are quantized using 8b fixed-point.}%
    \label{fig:vww}%
\end{figure}

As in \cite{chowdhery2019visual}, we employ the modified MobileNetV1 architecture which has a FC layer with 2 output classes instead of 1000.  The complexity of the network is tuned by varying the network width multiplier\cite{howard2017mobilenets} $m\in \{0.125,0.25,0.375,0.5\}$. Similar to our ImageNet experiments, we quantize all layers to 8b fixed-point and vary the precision of the PW layers using 8b-to-2b fixed-point and DBQ-1T and DBQ-2T.

As shown in Fig.~\ref{fig:vww-cc}, for over parameterized models, e.g., $m=0.5$, we find DBQ-1T  (red square) shows a massive reduction in $\mathcal{C}_S$ ($\sim 69\%$) at iso-accuracy compared to the fixed-point models (red circle) (Fig.~\ref{fig:vww-cc}). In contrast, for lightweight models, e.g., $m=0.125$, DBQ-1T (blue square) achieves an impressive $45\%$ reduction in $\mathcal{C}_S$ but at the expense of a $3\%$ loss in test accuracy as compared to the fixed-point model (blue circle) (Fig.~\ref{fig:vww-cc}). The DBQ models (diamonds and squares) can be seen to form a pareto-optimal accuracy-vs. $\calC_S$ trade-off curve in  Fig.~\ref{fig:vww-cc} demonstrating its effectiveness. 

Fig.~\ref{fig:vww-rc} shows that the choice of the width multiplier $m$ has a much more significant impact on the representational cost $\calC_R$ than varying bit-precision which implies that $\calC_R$ is dominated by the storage requirements of activations rather than weights. This implies that the choice of the model parameter $m$ is governed by the amount of on-chip storage available on an Edge device. In contrast, the choice of the bit precision of the PW layers is dictated by the latency/energy requirements which upper bounds $\mathcal{C}_S$ as seen in Fig.~\ref{fig:vww-cc}. As a result, when comparing the lightweight $m=0.25$ DBQ-2T model (orange diamond) with the over parameterized $m=0.375$ DBQ-1T model (green square), we observe that DBQ-2T achieves a reduction in both $\calC_S$ ($26\%$) and $\calC_R$ ($30\%$), at iso-accuracy and iso-$\calC_M$ ($\sim 10^6$b).

\section{Conclusion}
We presented DBQ, an efficient fully differentiable method for training multiple ternary branch quantizers for deep neural networks and validated its effectiveness for lightweight networks on the CIFAR-10 (ResNet-20), ImageNet (MobileNetV1, MobileNetV2, and ShuffleNetV2) and Visual Wake Words (MobileNetV1) datasets. Our method outperforms the state-of-the-art quantization schemes in both accuracy and complexity metrics.\\
\\
\textbf{Acknowledgment:} The authors would like to thank Avishek Biswas, Manu Mathew and Arthur Redfern for helpful discussions and support.

%
%
\bibliographystyle{splncs04}
\bibliography{egbib}

\includepdf[pages=-]{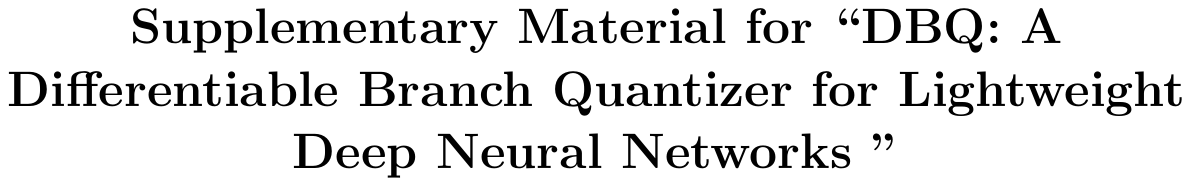}
\end{document}


\pagestyle{headings}
\mainmatter
\def\ECCVSubNumber{5625}  

\title{Supplementary Material for ``DBQ: A Differentiable Branch Quantizer for Lightweight Deep Neural Networks ''} 

\titlerunning{A Differentiable Branch Quantizer for Lightweight Deep Neural Networks}
%
\author{ }
\authorrunning{H. Dbouk et al.}
%
\institute{ }

\maketitle
\tableofcontents
\newpage
\section{Experimental Setup}
In this section, we describe the experimental setup used for generating all our results.
\subsection{CIFAR-10}
\subsubsection{Data Augmentation}
The CIFAR-10 dataset consits of $32\times 32$ RGB images. 
For generating the training samples, we adopt the standard data augmentation used in \cite{huang2018condensenet} where each image is: 1) zero-padded with $4$ pixels on each side; 2) horizontally flipped with probability $0.5$; and 3) randomly cropped using a $32\times 32$ window. During testing, we use the $32\times 32$ images as is from the testing set. 
We also normalize the images, for both training and testing, using a per-channel mean and standard deviation calculated across the training set. 
\subsubsection{Training Hyperparameters}
For training the full precision (FP) ResNet-20 baseline on CIFAR-10, we use SGD with momentum $\beta=0.9$, batch size of $100$, and weight decay of $\lambda=10^{-4}$. The FP model is trained for a total of $E_\text{T}=200$ epochs, with an initial learning rate $\eta_0=0.1$ and a cosine update rule \cite{loshchilov2016sgdr}:
\begin{equation}\label{eq:cosine}
    \eta_e = \frac{\eta_0}{2}\Big(1+\cos{\Big(\frac{ e}{E_{\text{T}}}\pi\Big)}\Big)
\end{equation}
During the fine-tuning process, i.e. training the model with weights initialized from the FP baseline, we train using the same setup as before, but for a fewer number of epochs $E_\text{T}=50$ and a smaller initial learning rate $\eta_0=0.01$. The DBQ models trained use a linear temperature increment schedule:
\begin{equation}
    T_e = T_{\text{init}} + e\cdot T_{\text{inc}}
\end{equation}
with an initial temperature $T_{\text{init}}=5$ and increments $T_{\text{inc}}=2.5$.

\subsection{ImageNet}
\subsubsection{Data Augmentation}
For our ImageNet experiments, we follow the standard data augmentation used in \cite{he2016deep}, where during training, images are: 1) resized; 2) horizontally flipped; and 3) randomly cropped to $224\times 224$. During testing, all images are resized to $256\times 256$ and then cropped to $224\times 224$. We also normalize the input images on a per-channel basis. 
\subsubsection{Training Hyperparameters}
For training the full precision MobileNetV1 baseline on ImageNet, we use a similar setup as our CIFAR-10 experiments, with a slightly different learning rate schedule. Similar to \cite{goyal2017accurate}, the first $E_{\text{W}}$ epochs are used for learning rate "warm-up":
\begin{equation}
    \eta_e = \frac{(e+1)\eta_0}{E_{W}}
\end{equation}
after which the remaining epochs utilize a cosine learning rate as described in \eqref{eq:cosine}. The hyperparameters used for both FP and quantization fine-tuning are specified in Table~\ref{tab:imagenet-setup-mnv1}.

The full precision MobileNetV2 and ShuffleNetV2 baselines on ImageNet are pre-trained models obtained from PyTorch \cite{paszke2017automatic}. Their 2T quantized counterparts, MobileNetV2-2T and ShuffleNetV2-2T, are fine-tuned using the training hyperparameters described in Table~\ref{tab:imagenet-setup-rest}.

\begin{table}[htbp]
    \begin{center}
    \begin{tabular}{|c||c|c|c|c|c|c||c|c|}
   \hline
    & Batch Size & $\beta$& $\lambda$ & $\eta_0$ & $E_{\text{W}}$ & $E_{\text{T}}$ & $T_{\text{init}}$ & $T_{\text{inc}}$\\ \hline \hline
    FP & $512$ & $0.9$ & $4\times10^{-5}$& $0.1$ & $5$ & $150$ & NA & NA \\ \hline
    Quant. & $512$ & $0.9$ & $4\times10^{-5}$& $0.001$ & $0$ & $50$ & $50$ & $20$ \\ \hline
    \end{tabular}
    \end{center}
    \caption{Training hyperparameters used for MobileNetV1 experiments on the ImageNet dataset.}
    \label{tab:imagenet-setup-mnv1}
\end{table}

\begin{table}[htbp]
    \begin{center}
    \begin{tabular}{|c||c|c|c|c|c|c||c|c|}
   \hline
    & Batch Size & $\beta$& $\lambda$ & $\eta_0$ & $E_{\text{W}}$ & $E_{\text{T}}$ & $T_{\text{init}}$ & $T_{\text{inc}}$\\ \hline \hline
    MobileNetV2-2T & $256$ & $0.9$ & $4\times10^{-5}$& $5\times10^{-4}$ & $0$ & $50$ & $25$ & $10$ \\ \hline
    ShuffleNetV2-2T & $512$ & $0.9$ & $4\times10^{-5}$& $0.001$ & $0$ & $30$ & $25$ & $10$ \\ \hline
    \end{tabular}
    \end{center}
    \caption{Training hyperparameters used for quantized MobileNetV2 and ShuffleNetV2 experiments on the ImageNet dataset.}
    \label{tab:imagenet-setup-rest}
\end{table}
\subsection{Visual Wake Words}
\subsubsection{Data Augmentation}
For data augmentation during training, we follow the exact setup as our ImageNet experiments with input normalization and random horizontal flips and crops. During testing, images are normalized, resized to $256\times 256$, and then cropped to $224\times 224$.
\subsubsection{Training Hyperparameters}
The training setup used is identical to our ImageNet experiments as well, and Table~\ref{tab:vww-setup} specifies the values of the hyperparameters used for both full precision and quantization training.

\begin{table}[htbp]
    \begin{center}
    \begin{tabular}{|c||c|c|c|c|c|c||c|c|}
   \hline
    & Batch Size & $\beta$& $\lambda$ & $\eta_0$ & $E_{\text{W}}$ & $E_{\text{T}}$ & $T_{\text{init}}$ & $T_{\text{inc}}$\\ \hline \hline
    FP & $512$ & $0.9$ & $4\times10^{-5}$& $0.1$ & $5$ & $200$ & NA & NA \\ \hline
    Quant. & $512$ & $0.9$ & $4\times10^{-5}$& $0.01$ & $0$ & $50$ & $20$ & $5$ \\ \hline
    \end{tabular}
    \end{center}
    \caption{Training hyperparameters used for experiments on the Visual Wake Words dataset.}
    \label{tab:vww-setup}
\end{table}

\section{Gradient Derivations}
In this section, we provide derivations for the gradient expressions of the loss function $\mathcal{L}$ with respect to the full precision weights $\mathbf{w} \in \reals^D$ and the quantizer parameters $\mathcal{P}_Q=\{\alpha_1, ..., \alpha_B, \gamma_1, \gamma_2, t_1, ..., t_{N-1}\}$. Recall that during training, the quantizer expression is:
\begin{equation}\label{eq:quantizer}
    \mathbf{z} = Q_T(\mathbf{w}) = \gamma_2 \Bigg[\sum_{i=1}^{N-1}\Big[\hat{f}_T(\gamma_1 \mathbf{w}-t_i)\sum_{j=1}^Bb_{i,j}\alpha_j\Big]-\sum_{j=1}^B\alpha_j\Bigg]
\end{equation}
where $\hat{f}_T$ is the smooth approximation using the Sigmoid function: 
\begin{equation}
   \hat{f}_T(u) = \frac{1}{1+\text{exp}(-Tu)} 
\end{equation}
whose derivative can be easily written as:
\begin{equation} \label{eq:sig-grad}
    \grad{\hat{f}_T(u)}{u} = T \hat{f}_T(u)\Big[1-\hat{f}_T(u)\Big] 
\end{equation}

\subsection{Notation}
The derivations of these gradients involves computing derivatives with vectors. Thus, in this section we establish the appropriate notation.
The derivative of a scalar $y$ with respect to a $D$-dimensional vector $\mathbf{x}$ is:
\begin{equation}\label{eq:d1}
    \frac{\partial y}{\partial\mathbf{x}} = \begin{bmatrix} \frac{\partial y}{\partial x_1} & \frac{\partial y}{\partial x_2} & \dots & \frac{\partial y}{\partial x_D} \end{bmatrix}
\end{equation}
whereas the derivative of a vector $\mathbf{y}$ with respect to a scalar $x$ is:
\begin{equation}\label{eq:d2}
    \frac{\partial \mathbf{y}}{\partial x} = \begin{bmatrix} \frac{\partial y_1}{\partial x} \\ \frac{\partial y_2}{\partial x} \\ \vdots \\ \frac{\partial y_D}{\partial x} \end{bmatrix}
\end{equation}
The derivative of a scalar $y$ with respect to another scalar $x$, assuming $y=g(\mathbf{z})$ and $\mathbf{z} = f(x)$, can therefore be computed using the chain rule:
\begin{equation}
    \grad{y}{x}=\grad{y}{\mathbf{z}}\cdot \grad{\mathbf{z}}{x} = \begin{bmatrix} \frac{\partial y}{\partial z_1} & \frac{\partial y}{\partial z_2} & \dots & \frac{\partial y}{\partial z_D} \end{bmatrix} \begin{bmatrix} \frac{\partial z_1}{\partial x} \\ \frac{\partial z_2}{\partial x} \\ \vdots \\ \frac{\partial z_D}{\partial x} \end{bmatrix} = \sum_{k=1}^D\grad{y}{z_k}\cdot \grad{z_k}{x}
\end{equation}

\subsection{Derivations}
\subsubsection{Post-quantization Scale}
We notice that:
\begin{equation}
    \grad{z_k}{\gamma_2} = \frac{z_k}{\gamma_2}
\end{equation}
which can be plugged in to get the gradient using the chain rule:
\begin{equation}
    \frac{\partial \calL}{\partial \gamma_2} = \frac{\partial \calL}{\partial \mathbf{z}} \cdot \frac{\partial \mathbf{z}}{\partial \gamma_2} = \sum_{k=1}^D \grad{\calL}{z_k}\cdot \grad{z_k}{\gamma_2} =  \frac{1}{\gamma_2} \sum_{k=1}^D\frac{\partial \calL}{\partial z_k}z_k
\end{equation}
\subsubsection{Ternary Branch Scales}
We first compute $\forall j \in [B]$:
\begin{equation}
    \grad{z_k}{\alpha_j} = \gamma_2 \Bigg[\sum_{i=1}^{N-1}\Big[\hat{f}_T(\gamma_1 w_k-t_i)b_{i,j}\Big]-1\Bigg] = \gamma_2 \Bigg[\sum_{i=1}^{N-1}\Big[g_{k,i}b_{i,j}\Big]-1\Bigg]
\end{equation}
where $g_{k,i}=\hat{f}_T(\gamma_1 w_k-t_i)$ for brevity. Therefore, using the chain rule we obtain:
\begin{equation}
    \grad{\calL}{\alpha_j} =  \grad{\calL}{\mathbf{z}}\cdot \grad{\mathbf{z}}{\alpha_j} =\sum_{k=1}^D \grad{\calL}{z_k}\cdot \grad{z_k}{\alpha_j} =  \gamma_2 \sum_{k=1}^D \frac{\partial \calL}{\partial z_k} \Bigg[\sum_{i=1}^{N-1}\Big[b_{i,j}g_{k,i}\Big]-1\Bigg]
\end{equation}
\subsubsection{Quantizer Thresholds}
We first utilize \eqref{eq:sig-grad} in order to compute $\forall i \in [N-1]$:
\begin{align}
    \begin{split}
     \grad{z_k}{t_i} &= \gamma_2 \Big[\grad{\hat{f}_T(\gamma_1 w_k-t_i)}{t_i}\sum_{j=1}^Bb_{i,j}\alpha_j\Big] \\
     &= -\gamma_2 T \Big[g_{k,i}(1-g_{k,i})\sum_{j=1}^Bb_{i,j}\alpha_j\Big] = -\gamma_2 T \Big[h_{k,i}\sum_{j=1}^Bb_{i,j}\alpha_j\Big]
    \end{split}
\end{align}
where $h_{k,i} = g_{k,i}(1-g_{k,i})$ for brevity. Therefore using the chain rule we obtain:
\begin{equation}
    \grad{\calL}{t_i} = \grad{\calL}{\mathbf{z}}\cdot \grad{\mathbf{z}}{t_i} =\sum_{k=1}^D \grad{\calL}{z_k}\cdot \grad{z_k}{t_i} = -\gamma_2 T \sum_{k=1}^D \frac{\partial \calL}{\partial z_k} \Big[h_{k,i}\sum_{j=1}^Bb_{i,j}\alpha_j \Big] 
\end{equation}
\subsubsection{Pre-quantization Scale}
Similarly, we utilize \eqref{eq:sig-grad} in order to compute:
\begin{align}
    \begin{split}
        \grad{z_k}{\gamma_1} &= \gamma_2 \Bigg[\sum_{i=1}^{N-1}\Big[\grad{\hat{f}_T(\gamma_1 w_k-t_i)}{\gamma_1}\sum_{j=1}^Bb_{i,j}\alpha_j\Big]\Bigg]= \gamma_2 T w_k \Bigg[\sum_{i=1}^{N-1}\Big[h_{k,i}\sum_{j=1}^Bb_{i,j}\alpha_j\Big]\Bigg]
    \end{split}
\end{align}
and therefore applying the chain rule yields:
\begin{equation}
    \grad{\calL}{\gamma_1} = \grad{\calL}{\mathbf{z}}\cdot \grad{\mathbf{z}}{\gamma_1} = \sum_{k=1}^D \grad{\calL}{z_k}\cdot \grad{z_k}{\gamma_1} = \gamma_2T \sum_{k=1}^D \grad{\calL}{z_k} w_k \Bigg[\sum_{i=1}^{N-1}\Big[h_{k,i}\sum_{j=1}^Bb_{i,j}\alpha_j\Big]\Bigg]
\end{equation}

\subsubsection{Full Precision Weights}
Finally, in order to compute the gradient of $\mathcal{L}$ with respect to the full precision weights $\mathbf{w}=[w_1, ..., w_D]^{\text{T}}$, we first compute $\forall k \in [D]$: 
\begin{align}
    \begin{split}
     \grad{z_m}{w_k} &= \gamma_2 \Bigg[\sum_{i=1}^{N-1}\Big[\grad{\hat{f}_T(\gamma_1 w_m-t_i)}{w_k}\sum_{j=1}^Bb_{i,j}\alpha_j\Big]\Bigg] \\
     &= \begin{cases} \gamma_1 \gamma_2 T \Bigg[\sum_{i=1}^{N-1}\Big[h_{k,i}\sum_{j=1}^Bb_{i,j}\alpha_j\Big]\Bigg], & \text{if } m=k  \\ 0, & \text{otherwise} \end{cases}
    \end{split}
\end{align}
and using the chain rule, we obtain:
\begin{equation}
    \grad{\calL}{w_k}= \grad{\calL}{\mathbf{z}}\cdot \grad{\mathbf{z}}{w_k} = \sum_{m=1}^D\grad{\calL}{z_m}\cdot \grad{z_m}{w_k} = \gamma_1 \gamma_2 T  \frac{\partial \calL}{\partial z_k} \sum_{i=1}^{N-1} \Big[ h_{k,i}\sum_{j=1}^Bb_{i,j}\alpha_j \Big]
\end{equation}

\section{MobileNetV2 on ImageNet Comparisons}
We compare DBQ and \cite{Uhlich2020Mixed} on MobileNetV2 in Table~\ref{tab:mn2-extra-results}. \cite{Uhlich2020Mixed} has two versions trained models M1 and M2, where M1 is trained with a memory constraint and M2 is not. We find that DBQ-2T is smaller than M2 \cite{Uhlich2020Mixed} at iso-accuracy on ImageNet and more accurate than M1 \cite{Uhlich2020Mixed} but at a larger storage cost. We are unable to compare the computational complexities since \cite{Uhlich2020Mixed} lacks sufficient information, hence we adopt the metrics reported in \cite{Uhlich2020Mixed}, which are weight storage (analogous to $\calC_M$) and activation storage (analogous to $\calC_R - \calC_M$).
\begin{table}[hbb]
    \begin{center}
    \resizebox{\columnwidth}{!}{%
    \renewcommand{\arraystretch}{1.2}
    \begin{tabular}{l|c c c}
    \clineB{1-4}{2.5}
    \textbf{Model} &  \textbf{Top-1 Acc.} [$\%$] & \textbf{Weight Storage} [MB] &  \textbf{Activation Storage} [MB]\\ \clineB{1-4}{2.5}
    M1 \cite{Uhlich2020Mixed} (w/ constr.) & $69.74$ & $1.55$ & $0.57$ \\ \hline
    M2 \cite{Uhlich2020Mixed} (w/o constr.)  & $70.59$ & $3.14$ & $1.58$ \\ \hline
    DBQ-2T & $\mathbf{70.54}$ & $\mathbf{2.43}$ & $\mathbf{1.15}$ \\ \hline 
    \end{tabular}
    }
    \end{center}
    \caption{The Top-1 accuracy on ImageNet and Storage costs for MobileNetV2 using our method (DBQ-2T) compared to \cite{Uhlich2020Mixed}.}
    \label{tab:mn2-extra-results}
\end{table}

\section{DBQ Branch Sparsity}

\begin{table}[htbp]
    \begin{center}
    \begin{tabular}{|c||c|c|c|c|c|}
    \cline{4-6}
    \multicolumn{3}{c}{}&\multicolumn{3}{|c|}{Average Branch Sparsity [$\%$]}\\
    \hline
    PW Layer & $C_{\text{in}}$ &  $C_{\text{out}}$ & FX8& DBQ-1T & DBQ-2T\\ \hline \hline
    0 & $64$ & $32$ & $35.55$& $58.69$ & $64.82$\\ \hline
    1 & $64$ & $128$ & $10.74$ & $41.42$& $51.75$\\ \hline
    2 & $128$ & $128$ & $6.86$ & $34.09$ & $46.45$\\ \hline
    3 & $128$ & $256$ & $6.73$ & $31.83$& $44.96$\\ \hline
    4 & $256$ & $256$ & $4.53$ & $29.10$& $43.05$\\ \hline
    5 & $256$ & $512$ & $7.31$ & $30.62$& $44.36$\\ \hline
    6 & $512$ & $512$ & $6.41$ & $28.50$& $43.40$\\ \hline
    7 & $512$ & $512$ & $6.00$ & $26.48$& $42.94$\\ \hline
    8 & $512$ & $512$ & $4.00$ & $24.03$& $41.70$\\ \hline
    9 & $512$ & $512$ & $5.57$ & $24.89$& $42.56$\\ \hline
   10 & $512$ & $512$ & $5.50$ & $23.65$& $42.30$\\ \hline
   11 & $512$ & $1024$ & $7.00$ & $23.17$& $42.41$\\ \hline
   12 & $1024$ & $1024$ & $10.69$ & $28.25$&$45.77$ \\ \hline \hline
   \multicolumn{3}{|c|}{Network Average} &$7.59$ &$26.50$& $\mathbf{43.78}$ \\\hline
    \end{tabular}
    \end{center}
    \caption{Branch level sparsity for all the pointwise (PW) layers of MobileNetV1 on ImageNet. $C_{\text{in}}$ and $C_{\text{out}}$ denote the number of input and output channels respectively.}
    \label{tab:sparsity}
\end{table}

One of the advantages of implementing ternary-based dot products is leveraging weight sparsity, which is reflected in our sparsity-aware computational cost $\calC_S$. In this work, we show that for MobileNetV1 on ImageNet with two ternary branch quantization (DBQ-2T-4), the computational cost can be reduced from $2.18\times 10^{10}$ FAs to $1.42\times 10^{10}$ ($\sim 35\%$ reduction) by simply skipping the operations involving zero weights. Table \ref{tab:sparsity} reports the average branch level sparsity for every point wise layer. For the DBQ-2T model, which quantizes PW layers to two ternary branches, we find that on average $43.78\%$ of all PW weights are zero, which explains the massive $35\%$ reduction in $\calC_S$. In contrast, the DBQ-1T model, which quantizes all PW layers to one ternary branch, achieves a $26.5\%$ average branch sparsity. While DBQ-2T has twice the number of branches compared to DBQ-1T, the per-branch sparsity is actually much higher for the DBQ-2T. In other words, while the number of pointwise parameters increases by $2\times$ when going from 1T to 2T, due to the high branch sparsity, the number of non-zero parameters increases by $1.53\times$ only. On the other hand, using 8b fixed-point for the PW layers yields very little weight sparsity ($7.59\%$). 

%
%
\bibliographystyle{splncs04}
\bibliography{egbib}